\documentclass[11pt]{article}

\usepackage[preprint]{acl}

\usepackage{times}
\usepackage{latexsym}
\usepackage{enumitem}
\usepackage{tikz}
\usepackage{algorithm}
\usepackage{algpseudocode}
\usetikzlibrary{positioning, arrows.meta, shapes.geometric, fit,backgrounds}
\usepackage{natbib}

\usepackage[T1]{fontenc}
\usepackage[utf8]{inputenc}

\usepackage{microtype}

\usepackage{inconsolata}

\usepackage{graphicx}
\usepackage{booktabs}
\usepackage{amsmath}
\usepackage{multirow}
\usepackage{subcaption}
\usepackage[table]{xcolor}
\title{ModularSQL: A Runtime Guardrail for the Multiplicity Blind Spot in Text-to-SQL}

\author{Tianxin Zhou \\
  University of Southern California\\
  \texttt{zhoutx0@gmail.com} \\\And
  Ruixi Lin \\
  Northeastern University\\
  \texttt{lin.ruix@northeastern.edu} \\}

\begin{document}
\raggedbottom
\maketitle
\begin{abstract}
Text-to-SQL systems are increasingly deployed atop production-style databases, where queries that pass benchmark evaluation can still produce results that distort downstream workflows. A query missing \texttt{DISTINCT} can inflate downstream aggregates such as \texttt{SUM} and report incorrect totals; duplicated rows can pollute analytics dashboards; a Cartesian-style join explosion can trigger query timeouts and cost overruns. Standard set-based execution accuracy (Set-EX) implicitly collapses duplicate rows in execution outputs and therefore fails to surface these deployment risks.

We call this the \textbf{Multiplicity Blind Spot} (MBS). To quantify it, we operationalize \textbf{Multiset-EX}, a multiplicity-preserving criterion that retains duplicate row counts. Across released DeepEye-SQL prediction artifacts from three open-source backbones: Qwen2.5-Coder-32B, Qwen3-Coder-30B-A3B, and Gemma-3-27B, on the executable BIRD-Dev subset ($N=1{,}532$), we observe a stable Set-EX vs.\ Multiset-EX gap of 5.81--6.79\,pp, suggesting that production-relevant multiplicity errors recur even in released Text-to-SQL artifacts. The gap persists beyond this pipeline family: released DAIL-SQL+GPT-4 predictions show a 5.22\,pp gap and the official BIRD GPT-3.5-turbo baseline a 3.39\,pp gap on the same subset.

We then introduce \textbf{ModularSQL}, a post-selection runtime guardrail. ModularSQL probes each executed result for multiplicity anomalies and applies deterministic patches or low-cost LLM rescues only on flagged queries, leaving unaffected queries unchanged. Integrated atop the DeepEye-SQL reproduction with Qwen3-Coder, ModularSQL preserves the Set-EX baseline (72.06\%), improves Multiset-EX from 65.86\% to 67.75\% (+1.89\,pp), and flags 77 high-risk anomalies, predominantly cardinality explosions, while adding only \$0.0076 in total LLM cost and 120\,ms amortized latency per query in our run. These results suggest that benchmark execution accuracy does not necessarily imply execution-safe SQL and that lightweight multiplicity-aware guardrails can narrow this gap with low deployment overhead.

Code will be available at \url{https://github.com/Ruixi1313/ModularSQL}

\end{abstract}

\section{Introduction}
Text-to-SQL is increasingly being used as an interface to databases in production settings for business intelligence, analytics, and decision support. However, SQL that is correct under benchmark evaluation does not necessarily guarantee execution safety. When duplicate tuples are incorrectly introduced or removed, the resulting query can distort aggregates, dashboards, and downstream analytical workflows. This gap between success and execution safety motivates our study.

LLM-based Text-to-SQL systems increasingly rely on multi-stage pipelines involving schema linking, value grounding, execution feedback, revision, and candidate selection \citep{li2026deepeye}, and are commonly evaluated on benchmarks such as BIRD \citep{li2023bird} and Spider \citep{yu2018spider}. Standard execution accuracy considers a predicted query correct if its result agrees with the gold query on a given database. Evaluation protocols differ in how matches are defined: BIRD’s official evaluator treats results as \emph{sets}, collapsing duplicate rows, whereas Spider’s test-suite evaluator treats them as multisets \citep{zhong2020test}. We therefore follow the set-based protocol used for all BIRD leaderboard results. This evaluation tolerates surface-form variation, but, as previous work shows, only approximates semantic correctness \citep{zhong2020test} and can produce false positives or negatives \citep{kim2025flex}. 

Existing critiques primarily focus on semantic equivalence: whether two SQL programs express the same logical intent. We identify a complementary gap: execution correctness under set-based comparison, which we denote as \textbf{Set-EX}, does not imply execution safety. Classical relational algebra treats relations as sets \citep{codd1970relational}, and many Text-to-SQL evaluators similarly compare output after collapsing duplicate tuples. This collapse discards multiplicity information that remains observable in actual SQL execution, where duplicates persist unless explicitly removed by \texttt{DISTINCT} or aggregation \citep{dayal1982duplicate}. A predicted query may therefore receive full execution credit while producing duplicated rows through a missing \texttt{DISTINCT}, an erroneous join, or a Cartesian-style explosion, the same failure modes that distort production analytics. 

We call this phenomenon the \textbf{Multiplicity Blind Spot} (MBS): the evaluator discards row multiplicity before comparison, so a Text-to-SQL system can appear correct under conventional execution accuracy while remaining unsafe for realistic deployment. To measure this gap, we use \textbf{Multiset-EX} as a complementary diagnostic criterion that preserves tuple multiplicities during result comparison. We do not propose it as a replacement for official benchmark metrics. Instead, it allows us to quantify failures that Set-EX cannot observe. On the executable BIRD-Dev subset ($N=1{,}532$), released DeepEye-SQL artifacts from Qwen2.5-Coder-32B,  Qwen3-Coder-30B-A3B, and Gemma-3-27B show a stable 5.81--6.79\,pp Set-EX versus Multiset-EX gap. The gap is not unique to this pipeline family: released DAIL-SQL+GPT-4 predictions show a 5.22,pp gap, while the official BIRD GPT-3.5-turbo baseline shows a 3.39,pp gap on the same query set.

We then introduce \textbf{ModularSQL}, a post-selection runtime guardrail that operates atop existing pipelines. ModularSQL probes
each executed result for multiplicity anomalies (e.g.,
duplication-ratio thresholds) and applies deterministic patches or
low-cost LLM rescues only on flagged queries, leaving unaffected
queries unchanged. Integrated atop our DeepEye-SQL reproduction with
Qwen3-Coder, ModularSQL preserves the Set-EX baseline (72.06\%). Under the stricter Multiset-EX metric, the same reproduced baseline scores 65.86\%, and ModularSQL improves it to 67.75\% (+1.89\,pp). It also flags 77 high-risk anomalies, predominantly cardinality explosions, while adding only \$0.0076 to the total LLM cost in our run. The candidate-free guardrail components transfer unchanged to two additional pipelines’ released predictions, improving Multiset-EX by 10 and 8 queries, respectively, while leaving Set-EX unchanged. These results suggest that benchmark execution accuracy does not necessarily imply execution-safe SQL. 

\noindent Our contributions are:
\begin{itemize}[noitemsep, topsep=0pt, partopsep=0pt, parsep=0pt]
    \item We identify the \textbf{Multiplicity Blind Spot} (MBS), a structural limitation of set-based execution evaluation that discards SQL multiplicity semantics and can mask multiplicity-sensitive execution risks.
    \item We operationalize \textbf{Multiset-EX} as a complementary multiplicity-aware diagnostic criterion to measure failures hidden by Set-EX.
    \item We provide cross-backbone \emph{and cross-pipeline} empirical evidence: three DeepEye-SQL backbone artifacts show a stable 5.81--6.79\,pp gap, and the gap persists on released DAIL-SQL+GPT-4 (5.22\,pp) and BIRD GPT-3.5-turbo (3.39\,pp) predictions.
    \item We introduce \textbf{ModularSQL}, a post-selection runtime guardrail that flags 77 high-risk multiplicity anomalies and improves Multiset-EX from 65.86\% to 67.75\% (+1.89\,pp) while preserving the reproduced Set-EX baseline at 72.06\%, with candidate-free components that transfer unchanged to two additional pipelines.
    \item We characterize the guardrail through detector precision/recall, threshold cross-validation, wall-clock latency, and a non-LLM selection baseline that isolates the multiplicity-aware signal from generic LLM repair ability.
\end{itemize}

\section{Background and Related Work}
\subsection{Text-to-SQL Pipelines and Execution Feedback}
\label{sec:pipelines}

Modern Text-to-SQL systems have evolved from direct SQL generation into multi-stage orchestration pipelines that decompose, regenerate, and select among candidate SQL queries based on intermediate execution feedback~\citep{pourreza2023dinsql, gao2024dailsql, li2026deepeye}. DeepEye-SQL~\citep{li2026deepeye}, the system we build on, follows a seven-stage workflow; we focus on its final three: \textbf{(S5)} \textit{candidate generation}, where the backbone LLM produces 12 candidate SQL queries per question; \textbf{(S6)} \textit{revision}, where candidates are iteratively repaired through LLM-driven rewrites guided by syntactic checks and execution errors; and \textbf{(S7)} \textit{selection}, where surviving candidates are adjudicated through a tournament procedure to produce the final SQL.

The execution feedback consulted in these stages is primarily \emph{validity-oriented}: it triggers on parse failures, invalid schema references, and runtime execution errors. Related LLM self-debugging work extends this feedback channel for general code generation~\citep{chen2024selfdebug}, but the multiplicity structure of executed output, including whether duplicates are sparse or dense and whether a join has amplified the row count, is not explicitly monitored by these feedback signals. This leaves a diagnostic gap that validity-oriented feedback alone does not close. 

\paragraph{Relation to repair and refinement methods.}
Execution-guided repair \citep{chen2024selfdebug} and multi-candidate refinement \citep{pourreza2023dinsql, gao2024dailsql, li2026deepeye} regenerate or revise SQL using execution feedback. ModularSQL is complementary: our main setting measures the 6.20,pp Multiset-EX gap \emph{after} DeepEye-SQL’s revision (S6) and tournament selection (S7). The guardrail closes 30.5\% of this residual gap post hoc while preserving Set-EX. We do not evaluate an additional full repair round; this limitation is stated in Section~\ref{sec:limitations}.

\subsection{From Set-EX to Multiplicity-Preserving Evaluation}
\label{sec:bagsemantics}

Let $q_p$ denote a predicted SQL query, $q_g$ the gold SQL query, and $D$ the target database. Let $R(q,D)$ be the \emph{multiset} of result rows produced by executing query $q$ on $D$. \textbf{Set-EX} treats a prediction as correct when the predicted and gold result rows match as sets. Let:
\[
\begin{aligned}
S_p &= \mathrm{set}(R(q_p,D)), \\
S_g &= \mathrm{set}(R(q_g,D)).
\end{aligned}
\]
\[
\mathrm{SetEX}(q_p,q_g,D)=\mathbf{1}[S_p=S_g].
\]
This robustly ignores row-order differences, but also collapses duplicate multiplicities. This set-collapsing behavior is not a simplification we introduce but follows the official BIRD protocol: both the main BIRD evaluator (\texttt{execute\_sql} in \texttt{evaluation.py})\footnote{\url{https://github.com/AlibabaResearch/DAMO-ConvAI/blob/main/bird/llm/src/evaluation.py}} and the BIRD Mini-Dev evaluator (\texttt{evaluation\_ex.py})\footnote{\url{https://github.com/bird-bench/mini_dev/blob/main/evaluation/evaluation_ex.py}} compare \texttt{set(predicted\_res) == set(ground\_truth\_res)}. In contrast, Spider’s test-suite evaluator uses multiset comparison (\texttt{multiset\_eq} in \texttt{exec\_eval.py}) \citep{zhong2020test}. Thus, MBS is a property of set-collapsing protocols such as BIRD’s official evaluation, not of Text-to-SQL evaluation in general. Previous critiques of execution-based evaluation focus on \emph{semantic equivalence}~\citep{zhong2020test, kim2025flex}, a direction orthogonal to ours: even when Set-EX certifies a prediction as answer-correct, its execution may still contain unsafe duplicate structure.

Classical relational algebra treats relations as sets~\citep{codd1970relational}, whereas actual SQL execution follows bag semantics: a \texttt{SELECT} without \texttt{DISTINCT} preserves duplicates that affect aggregation operators such as \texttt{COUNT} and \texttt{SUM}. \citet{dayal1982duplicate} formalize this through an extended algebra with explicit duplicate-elimination control.

Aligned with this perspective, we operationalize a
multiplicity-preserving variant of execution accuracy:
\[
\mathrm{MultisetEX}(q_p,q_g,D) =
\mathbf{1}\!\left[\, R(q_p,D) = R(q_g,D) \,\right],
\]
Here the comparison is order-insensitive but multiplicity-preserving: duplicate tuple counts are retained, while row order is ignored. Multiset-EX is at least as strict as Set-EX: $\mathrm{MultisetEX}=1 \Rightarrow \mathrm{SetEX}=1$, but the converse does not necessarily hold.

Queries that pass Set-EX while failing Multiset-EX realize the \textbf{Multiplicity Blind Spot} (MBS). In such cases, the $\mathrm{set}(\cdot)$ projection removes precisely the duplicate information that distinguishes the predicted and gold outputs, so Set-EX cannot detect the discrepancy by construction. This blindness is a property of the evaluator itself; its empirical incidence on any specific artifact depends on the producing pipeline. We use Multiset-EX as a complementary diagnostic lens, not as a replacement for Set-EX.

\section{ModularSQL Methodology}
\label{sec:method}
\begin{figure}[t]
\centering
\resizebox{\columnwidth}{!}{%
\begin{tikzpicture}[
  node distance=0.48cm,
  every node/.style={font=\small},
  box/.style={
    rectangle, draw=black!70, rounded corners=2pt,
    minimum width=5.2cm, minimum height=0.62cm,
    align=center, line width=0.45pt
  },
  pipeline/.style={
    box, fill=blue!8,
    minimum height=1.05cm
  },
  probe/.style={
    box, fill=orange!14,
    minimum height=1.05cm
  },
  branch/.style={
    rectangle, draw=black!70, rounded corners=2pt,
    minimum width=2.45cm, minimum height=0.78cm,
    align=center, line width=0.45pt
  },
  arr/.style={-{Latex[length=2mm,width=1.4mm]}, line width=0.55pt},
  dashedbox/.style={
    draw=black!45, dashed, rounded corners=3pt,
    inner sep=4pt
  }
]

\node[box, fill=gray!8] (input) {Question $+$ DB Schema};

\node[pipeline, below=of input] (deepeye) {%
  \textbf{DeepEye-SQL}\\[-1pt]
  \scriptsize S5 Candidate Gen. $\rightarrow$ S6 Revision $\rightarrow$ S7 Selection
};

\node[box, fill=gray!8, below=of deepeye] (selected) {Selected SQL};

\node[probe, below=of selected] (probe) {%
  \textbf{ModularSQL Runtime Probe}\\[-1pt]
  \scriptsize \texttt{dup\_ratio} / empty result / execution error
};

\node[branch, fill=green!10, below left=0.72cm and -0.05cm of probe]
  (normal) {Normal\\[-1pt]\scriptsize return unchanged};

\node[branch, fill=red!10, below right=0.72cm and -0.05cm of probe]
  (anom) {Anomalous\\[-1pt]\scriptsize patch or LLM rescue};

\node[box, fill=blue!8, below=1.15cm of probe] (final) {Final SQL};

\draw[arr] (input) -- (deepeye);
\draw[arr] (deepeye) -- (selected);
\draw[arr] (selected) -- (probe);
\draw[arr] (probe.south) -| (normal.north);
\draw[arr] (probe.south) -| (anom.north);
\draw[arr] (normal.south) |- (final.west);
\draw[arr] (anom.south) |- (final.east);

\begin{scope}[on background layer]
  \node[dashedbox, fit=(probe)(normal)(anom)(final)] (Modularbox) {};
\end{scope}

\node[anchor=west, font=\scriptsize\bfseries, text=black!65]
  at (Modularbox.north west) {ModularSQL post-selection guardrail};

\end{tikzpicture}%
}
\caption{Overview of ModularSQL. ModularSQL operates as a post-selection runtime guardrail layer atop DeepEye-SQL. After S7 selects an executable SQL query, ModularSQL probes the execution behavior for multiplicity-related anomalies such as high duplication ratio, empty results, or execution errors. Unaffected queries pass through unchanged, while anomalous queries are routed through targeted interventions, including deterministic patches or a low-cost LLM rescue.}
\label{fig:Modularsql-arch}
\end{figure}
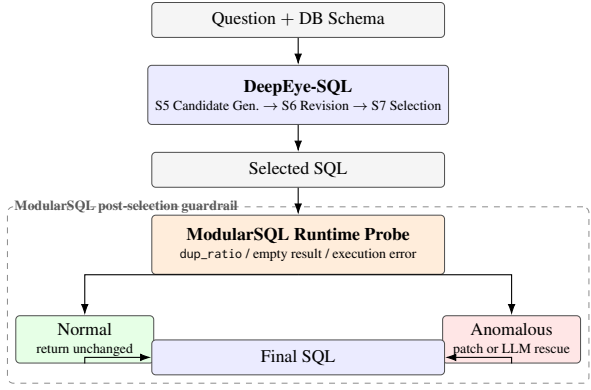

ModularSQL is a post-selection runtime guardrail layer that complements the DeepEye-SQL output rather than replacing the underlying pipeline. It probes the executed result of each selected SQL query for multiplicity-related anomalies and applies targeted interventions, including deterministic patches and low-cost LLM rescue only when an anomaly is detected. Figure~\ref{fig:Modularsql-arch} summarizes the architecture.

\subsection{Runtime Probe Design}
\label{sec:runtime-probe}

Given an executable SQL query $q_p$ produced by S7, the runtime probe attempts to execute $q_p$ on the target database $D$ and, when execution succeeds, obtains the result row multiset $R_p = R(q_p, D)$. The probe then inspects three health conditions:
(i)~execution error, (ii)~empty result, and (iii)~excessive duplicate multiplicity. Let $\mathrm{err}(q_p, D) \in \{0, 1\}$ be the indicator that executing $q_p$ on $D$ raises a database error. When $\mathrm{err}(q_p, D) = 0$ and $|R_p| > 0$, the duplication ratio
\begin{equation}
\mathrm{dup\_ratio}(R_p)
\;=\; \frac{|R_p| - |\mathrm{unique}(R_p)|}{|R_p|}
\;\in\; [0, 1)
\end{equation}
measures the fraction of result rows that are redundant duplicate copies, i.e., rows beyond the first occurrence of each distinct row. For compactness, let $\mathrm{err}=\mathrm{err}(q_p,D)$ and $\rho(R_p)=\mathrm{dup\_ratio}(R_p)$. A query is marked \emph{anomalous} by the following ordered rule:
\[
\mathrm{anomalous}(q_p) =
\begin{cases}
    1, & \mathrm{err}=1 \\
    1, & \mathrm{err}=0,\ |R_p| = 0, \\
    1, & \begin{aligned}[t]
        &\mathrm{err}=0,\ |R_p| > 0,\\
        &\rho(R_p)\geq \tau,
        \end{aligned}\\
    0 & \text{otherwise}.
    \end{cases}
\]

We deploy $\tau = 0.80$ as the operating point; the full sensitivity sweep and cross validation are reported in Section~\ref{sec:experiments}. 

The three conditions target different failure regimes. Condition~(i) catches queries that are demonstrably broken at the database level. Condition~(iii) targets the multiplicity-sensitive failures characterized in Section~\ref{sec:bagsemantics}: missing \texttt{DISTINCT}, join
amplification, and Cartesian-style row explosion, all of which may remain hidden under set-based evaluation. Condition~(ii) is a conservative trigger rather than a correctness claim: an empty result is sometimes legitimate (the gold query may also return no rows), but in deployment we flag it as a
probe signal that triggers downstream interventions (Sections~\ref{sec:patches} and~\ref{sec:rescue}). The multiplicity check requires a linear scan of $R_p$; conditions~(i) and~(ii) are read directly from the database response. No condition invokes an LLM. The probe targets \emph{runtime-observable physical anomalies}; semantic errors with normal-looking outputs are out of scope. The operating point prioritizes precision over recall (Section~\ref{sec:diagnostics}). Among the three conditions, condition~(i) overlaps with the validity-oriented feedback described in Section~\ref{sec:pipelines}, whereas conditions~(ii) and~(iii) introduce additional runtime signals: empty outputs and duplicate multiplicity anomalies.

The probe therefore induces an asymmetric intervention policy. Queries that do not trigger any anomaly are returned unchanged, while flagged queries are routed to targeted interventions. These interventions comprise a low-cost LLM rescue followed by deterministic patches (Sections~\ref{sec:rescue} and~\ref{sec:patches}); the ordering is empirically necessary and is examined in Section~\ref{sec:experiments}. This design preserves the behavior of the underlying Text-to-SQL pipeline on unflagged queries and keeps additional inference cost proportional to the number of anomalous cases rather than the total workload. In this sense, ModularSQL acts as a post-selection guardrail: it requires only the selected SQL and its executed result, rather than modifying candidate generation, revision, or selection.

\subsection{Deterministic Patches}
\label{sec:patches}

Following the LLM rescue layer (Section~\ref{sec:rescue}), ModularSQL applies two deterministic rewrites that target the most common multiplicity-sensitive failures. Both rewrites are syntactic and invoke no LLM. 

\textbf{P1 (DISTINCT injection).} If the duplication ratio of the predicted query's executed result satisfies \(\mathrm{dup\_ratio}(R_p) \geq 0.80\), ModularSQL inserts a \texttt{DISTINCT} clause into the outermost \texttt{SELECT}. When the query contains a \texttt{LIMIT} clause, \texttt{DISTINCT} only deduplicates before the \texttt{LIMIT} cut, so injection can alter even set-semantics results. We identified this guard through cross-pipeline evaluation (Section~\ref{sec:cross-pipeline}); none of the patch-eligible queries in our DeepEye-SQL reproduction contains \texttt{LIMIT}, so the reported results are unaffected. The threshold is equivalent to total rows being at least five times the number of distinct rows, the characteristic signature of join-induced cartesian explosion; below this regime, duplicates are more likely to be semantically meaningful (e.g., two distinct entities sharing a display name).

\textbf{P2 (DISTINCT removal).} For predicted queries that already contain a \texttt{DISTINCT} clause, ModularSQL tentatively removes it, re-executes the modified query, and checks whether the resulting row multiset has a very low duplication ratio (\(\mathrm{dup\_ratio} \leq 0.10\)). If so, the original \texttt{DISTINCT} is treated as a candidate over-application by the backbone model, and the rewritten query is committed.

The two thresholds are not symmetric: although a single split point between "too few" and "too many" duplicates might seem natural, empirically the duplication ratio is a strong signal for adding \texttt{DISTINCT} but only an actionable signal for removal in the extreme tail (\(\mathrm{dup\_ratio} \leq 0.10\)). This asymmetry suggests that unnecessary \texttt{DISTINCT}s introduced by LLMs reflect query-intent misclassifications rather than mechanical row amplification, and motivates treating the two patches as independent rules rather than as symmetric counterparts. Sensitivity to both thresholds is reported in Section~\ref{sec:experiments}.

\subsection{Budget-Aware LLM Rescue}
\label{sec:rescue}
For queries that the runtime probe flags as anomalous, ModularSQL invokes a low-cost LLM rescue step before the deterministic patches of Section~\ref{sec:patches}. The rescue treats the LLM as a judge that selects among the candidate set produced by the upstream pipeline rather than generating new SQL from scratch \citep{zheng2023llmjudge}. Algorithm~\ref{alg:rescue} summarizes the procedure.

\begin{algorithm}[!h]
\small
\caption{Budget-Aware LLM Rescue}
\label{alg:rescue}
\begin{algorithmic}[1]
\Require Predicted SQL $q_p$, candidate set $C$ from upstream stages, database $D$, previews $P$
\State $R_p \gets \textsc{Execute}(q_p,D)$
\Comment{Section~\ref{sec:runtime-probe}}
\If{$\mathrm{anomalous}(q_p)=0$}
    \State \Return $q_p$ \Comment{trust S7 selection}
\EndIf
\State $\textit{prompt} \gets
\textsc{BuildPrompt}(question, schema, q_p, C, P)$
\State $\textit{idx} \gets \textsc{LLMJudge}(\textit{prompt})$
\If{$\textit{idx} \geq 0$}
    \State \Return $C[\textit{idx}]$ \Comment{candidate replaces S7 output before patches}
\Else
    \State \Return $q_p$ \Comment{LLM declined; fall back to S7}
\EndIf
\end{algorithmic}
\end{algorithm}

\textbf{Input.} The LLM receives the natural-language question, a schema summary, the original S7 selection, the full set of revised candidates from S6 (twelve in our setting), and a short execution preview of each candidate (the first few result rows, an empty-result flag, or an execution error). It is also told which probe condition triggered the rescue. 

\textbf{Selection.} The LLM returns an index into the candidate set, or a special token indicating that no candidate looks safe. If a candidate is selected, ModularSQL replaces the S7 prediction with that candidate; if no selection is made, the S7 prediction is retained unchanged.

\textbf{Cost.} Because the rescue is gated by the probe, its LLM budget scales with the number of flagged queries rather than the total workload. On our BIRD-Dev reproduction the rescue fires on 77 of 1{,}532 queries and adds approximately \$0.0076 in total LLM cost in our run (under \$0.0001 per affected query under our provider pricing). End-to-end accuracy and per-condition trigger statistics are reported in Section~\ref{sec:experiments}.

\section{Experiments}
\label{sec:experiments}

\subsection{Experimental Setup}
\label{sec:setup}
We evaluate ModularSQL on the BIRD-Dev split \citep{li2023bird}, comprising 1{,}534 question-SQL-database triples. Gold and predicted SQL are executed under a 5\,s timeout, excluding two queries (qids 518, 701) whose gold SQL exceeds this limit and giving an evaluable subset of $N=1{,}532$. We reproduce the seven-stage DeepEye-SQL pipeline \citep{li2026deepeye} with Qwen3-Coder-30B-A3B \citep{qwen2025qwen3} as the backbone LLM. ModularSQL operates atop this reproduction with operating points $\tau=0.80$ for P1 (\texttt{DISTINCT} injection) and $\tau=0.10$ for P2 (\texttt{DISTINCT} removal). All predictions are scored under both Set-EX and Multiset-EX, defined in Section~\ref{sec:bagsemantics}. For cross-pipeline evaluation, we additionally use two released BIRD-Dev prediction sets: DAIL-SQL+GPT-4’s main configuration \citep{gao2024dailsql} and the official BIRD GPT-3.5-turbo (knowledge) baseline \citep{li2023bird}. As each provides a single final SQL per question, only the candidate-free detector and deterministic patches apply.

\subsection{Cross-Backbone Evidence of MBS}
\label{sec:cross-backbone}
We first establish that MBS is not specific to our reproduction by evaluating three released DeepEye-SQL prediction artifacts on the same evaluable BIRD-Dev subset ($N=1{,}532$). The artifacts are generated by Qwen2.5-Coder-32B \citep{hui2024qwen25coder}, Qwen3-Coder-30B-A3B \citep{qwen2025qwen3}, and Gemma-3-27B \citep{gemma2025gemma3}. Table~\ref{tab:cross-backbone} reports Set-EX and Multiset-EX scores for each artifact and for our own DeepEye-SQL reproduction with Qwen3-Coder, computed with the evaluator implementations described in Section~\ref{sec:bagsemantics}.

All four measurements exhibit a stable Set-EX versus Multiset-EX gap between 5.81 and 6.79 percentage points across different backbone families and released artifacts. Our reproduction (6.20\,pp) falls inside this range. This is consistent with our framing that the blindness is induced by set-based evaluation, while its empirical incidence recurs across artifacts, including those produced by multi-stage Text-to-SQL pipelines that already incorporate execution feedback and tournament selection.

\begin{table}[!h]
\centering
\scriptsize
\setlength{\tabcolsep}{8pt}
\begin{tabular}{lccc}
\toprule
Backbone & Set & Multi & Gap \\
\midrule
Qwen2.5-Coder-32B & 69.91 & 64.10 & 5.81 \\
Qwen3-Coder-30B-A3B & 72.85 & 66.51 & 6.33 \\
Gemma-3-27B & 70.50 & 63.71 & 6.79 \\
\midrule
\rowcolor{blue!8}Qwen3-Coder-30B-A3B (ours) & 72.06 & 65.86 & 6.20 \\
\bottomrule
\end{tabular}
\caption{Set-EX and Multiset-EX on executable BIRD-Dev ($N=1{,}532$).
Scores are percentages; gaps are percentage points.}
\label{tab:cross-backbone}
\end{table}

\subsection{Cross-Pipeline Transfer}
\label{sec:cross-pipeline}

To assess whether MBS and the candidate-free guardrail components transfer beyond DeepEye-SQL, we apply the detector and deterministic patches unchanged ($\tau=0.80$) to two released BIRD-Dev prediction sets (Section~\ref{sec:setup}). As shown in Table~\ref{tab:cross-pipeline}, the MBS gap persists (5.22\,pp and 3.39\,pp), while weaker pipelines trigger substantially more anomalies (270 and 555 vs.\ 77 for DeepEye-SQL). With the \texttt{LIMIT} guard (Section~\ref{sec:patches}), the deterministic patch improves Multiset-EX by $+10$ and $+8$ queries, respectively, with no change in Set-EX. We do not evaluate the candidate-based LLM rescue because these releases provide only one SQL per question and thus no candidate pool.
\begin{table}[!h]
\centering
\scriptsize
\setlength{\tabcolsep}{4pt}
\begin{tabular}{lccccc}
\toprule
System & Set-EX & Mset-EX & Gap & Flagged & Patched Mset \\
\midrule
DAIL-SQL+GPT-4 & 53.59 & 48.37 & 5.22 & 270 & 49.02 ($+10$) \\
BIRD GPT-3.5 (kg) & 35.90 & 32.51 & 3.39 & 555 & 33.03 ($+8$) \\
\midrule
DeepEye-SQL (ours) & 72.06 & 65.86 & 6.20 & 77 & --- \\
\bottomrule
\end{tabular}
\caption{Candidate-free guardrail components applied unchanged to two released BIRD-Dev prediction sets ($N{=}1{,}532$). Set-EX is unchanged by the patch on both systems.}
\label{tab:cross-pipeline}
\end{table}

\subsection{Main Results and Ablations}
\label{sec:main-results}

\begin{table}[!h]
\centering
\scriptsize
\setlength{\tabcolsep}{3pt}
\begin{tabular}{lccc}
\toprule
System & Set-EX (\%) & Multiset-EX (\%) & $\Delta$ Multiset-EX \\
\midrule
DeepEye reproduction                & 72.06 & 65.86 & --- \\
+ P1                                & 72.06 & 67.23 & $+1.37$ \\
+ P1 + P2                           & 72.06 & 67.43 & $+1.57$ \\
v1: P1+P2 $\rightarrow$ Rescue      & 72.06 & 67.17 & $+1.31$ \\
\rowcolor{blue!12} \textbf{v2: Rescue $\rightarrow$ P1+P2} & \textbf{72.06} & \textbf{67.75} & $\mathbf{+1.89}$ \\
v2 variant: dup-ratio selector & 72.19 & 67.30 & $+1.44$ \\
\bottomrule
\end{tabular}
\caption{Main results on the executable BIRD-Dev subset ($N=1{,}532$). $\Delta$ is measured relative to the DeepEye reproduction. Rows 1 to 3 give the cumulative effect of the deterministic patches; rows 4 to 5 contrast the two intervention orderings. The highlight v2 row is the canonical ModularSQL configuration used throughout the paper. The dup-ratio selector replaces the LLM rescue with a zero-cost rule that selects the healthy candidate with the lowest duplication ratio; unlike the LLM rescue, it does not fully preserve Set-EX, yielding a net gain of 2 Set-EX queries.}

\label{tab:main-results}
\end{table}

Table~\ref{tab:main-results} quantifies the end-to-end effect of each ModularSQL component on our DeepEye-SQL reproduction with Qwen3-Coder-30B-A3B. ModularSQL v2 preserves the Set-EX baseline (1{,}104 of 1{,}532; 72.06\%) and improves Multiset-EX from 65.86\% to 67.75\%, a net gain of 29 queries (+1.89\,pp). Set-EX is invariant across all five configurations while Multiset-EX improves over the baseline under every intervention, confirming that the guardrail changes execution outcomes specifically along the multiplicity dimension and leaves the underlying pipeline's
set-level behavior intact.

The two deterministic patches together account for 24 of the 29 net Multiset-EX rescues: P1 (DISTINCT injection at $\tau=0.80$) contributes 21 queries (+1.37\,pp), and P2 (DISTINCT removal at $\tau=0.10$) adds 3 more on top of P1 (+0.20\,pp). The remaining +5 queries in ModularSQL v2 come from the LLM rescue layer, which substitutes alternative S6 candidates before the patches refine the result.

Rows 4 and 5 of Table~\ref{tab:main-results} compare the two intervention orderings introduced in Section~\ref{sec:runtime-probe}. Applying the deterministic patches before the LLM rescue (v1) yields a Multiset-EX net gain of only +20 queries, 9 fewer than v2 (+29). The 0.58\,pp gap quantifies the cost of letting the LLM rescue overwrite already-patched SQL, and motivates the rescue-first ordering used throughout the rest of the paper.

\paragraph{Non-LLM rescue baseline.} Replacing the LLM rescue with a zero-cost selector that chooses the healthy candidate with the lowest duplication ratio yields 1{,}031/1{,}532 (67.30\%) Multiset-EX, versus 1{,}038 with the LLM rescue, recovering 22 of 29 net corrections (75.9\%). Thus, most gains come from the multiplicity-aware signal rather than generic LLM repair. The remaining difference isolates the LLM's role: among 76 evaluable flagged queries, 30 have no healthy candidate, causing the ratio rule to fall back to the base SQL; 12 of the LLM's 13 additional fixes occur in these cases.

\subsection{Analysis of Runtime Interventions}
\label{sec:analysis}

\paragraph{Trigger reasons.} Of the 77 queries the probe flags as anomalous, 70 (91\%) trigger condition (iii) (duplication ratio above $\tau$), 6 (8\%) trigger condition (ii) (empty result), and 1 triggers condition (i) (execution error). The duplication ratio signal therefore dominates trigger volume, consistent with our characterization of MBS as a multiplicity-amplification phenomenon. 
In our run, the LLM returned a valid candidate index for all 77 triggered cases; the fallback path defined in Algorithm~\ref{alg:rescue} was not invoked. 

\paragraph{Cost.} The intervention adds approximately \$0.0076 in LLM cost in our run (under \$0.0001 per affected query); the deterministic patches incur no inference cost. Because the LLM rescue is gated by the probe, end-to-end cost scales with the number of flagged queries rather than the total workload. Only 5.0\% (77/1{,}532) of evaluated queries trigger the rescue stage.

\paragraph{Rescue effect on flagged queries.} Within the 77 cases the probe flags, the LLM rescue improves Multiset-EX from 12/77 (15.6\%) to 29/77 (37.7\%) relative to the raw S7 selection, with 25 multiplicity-level fixes against 8 regressions. This local gain does not translate directly into end-to-end improvement, because the deterministic patches also modify some of the same queries. Consequently, the rescue layer contributes a net $+5$ queries to end-to-end Multiset-EX in v2. 

\paragraph{By-difficulty gain.} Table~\ref{tab:by-difficulty} breaks down the Multiset-EX gain by BIRD difficulty stratum. The largest gain is observed in the challenging stratum (+4.17\,pp), consistent with the intuition that complex multi-join queries expose more multiplicity-sensitive failure modes. Together, these results suggest that multiplicity-sensitive failures are relatively rare but disproportionately concentrated in complex queries, making them well suited for selective runtime intervention.

\begin{table}[!ht]
    \centering
    \scriptsize
    \setlength{\tabcolsep}{8pt}
    \begin{tabular}{lcccc}
    \toprule
    Difficulty & $n$ & Base & ModularSQL v2 & $\Delta$ \\
    \midrule
    Simple      & 925 & 71.57 & 73.19 & $+1.62$ \\
    Moderate    & 463 & 60.48 & 62.20 & $+1.73$ \\
    \rowcolor{blue!10} \textbf{Challenging} & \textbf{144} & \textbf{46.53} & \textbf{50.69} & $\mathbf{+4.17}$ \\
    \bottomrule
    \end{tabular}
    \caption{Multiset-EX gain by BIRD difficulty stratum ($N=1{,}532$). Scores are percentages; $\Delta$ is the ModularSQL v2 gain over the DeepEye reproduction. The challenging stratum shows the largest relative gain.}
    \label{tab:by-difficulty}
\end{table}

\subsection{Detector Diagnostics}
\label{sec:diagnostics}
With Multiset-EX failure as ground truth, the deployed trigger ($\tau=0.80$) is precision-first: precision is 0.844 on DeepEye-SQL, 0.945 on DAIL-SQL+GPT-4, and 0.991 on GPT-3.5-turbo. The exec-error and empty-result triggers achieve 1.000 precision, while the duplication trigger achieves 0.797--0.878 across systems. For MBS failures (Set-EX-correct but Multiset-EX-wrong), recall is 0.35--0.43. Recall over all SQL errors is necessarily partial because semantic errors with normal-looking outputs are invisible to runtime probing (Section~\ref{sec:runtime-probe}). Notably, the fixed $\tau$ transfers to unseen DAIL-SQL and GPT-3.5-turbo pipelines, with precision improving rather than degrading. Leave-one-database-out cross-validation over nine databases with P1-fire cases yields a held-out net gain of $+17$, non-negative in 8/9 folds (worst: $-2$); removing the threshold reduces the held-out net to $+5$. Full results are in Appendix~\ref{appx:diagnostics}.

\paragraph{Latency.} On a commodity laptop with local SQLite databases, detection reuses fetched result rows and costs 3\,$\mu$s median per query. The \texttt{DISTINCT} patch adds one execution on 56 queries (2\,ms median, 82\,ms p90). The LLM rescue, triggered on 77/1{,}532 queries, takes 1.48\,s median end-to-end (1.36\,s for the API call, 47\,ms for candidate re-execution). Overall, the guardrail adds 120\,ms per query amortized across the workload.

\section{Discussion}
\label{sec:discussion}

ModularSQL is designed as a deployment-friendly intervention layer that complements rather than replaces an existing Text-to-SQL pipeline. The deployment profile of this design appears clearly in our reproduction: the Set-EX score reported by BIRD-style evaluators is preserved exactly, the probe fires on only 5\% of queries, and the additional LLM cost over a 1{,}532-query workload is under one cent in our run, and the amortized latency overhead is 120\,ms per query. Deployment teams that already trust their selection-stage output can adopt the guardrail incrementally without retraining the backbone or modifying the upstream stages. Because the 6.20,pp gap remains after DeepEye-SQL’s own revision and selection stages, the guardrail complements rather than replaces existing repair and refinement methods.

The observed 30.5\% reduction in the MBS gap also highlights a broader evaluation consideration. Set-EX, while operationally useful for benchmarking, is structurally blind to multiplicity-level errors that real downstream applications, such as analytics dashboards, aggregate reports, and materialized views, cannot tolerate. Pairing benchmark Set-EX with a complementary multiplicity-aware diagnostic such as Multiset-EX can help surface this gap during evaluation rather than at deployment. The residual 4.31\,pp gap suggests that multiplicity-sensitive failures are not exhausted by duplicate-elimination patterns alone. 

\section{Conclusion}
\label{sec:conclusion}

We introduced ModularSQL, a post-selection runtime guardrail that exposes and partially mitigates the Multiplicity Blind Spot in benchmark-correct Text-to-SQL pipelines. On our DeepEye-SQL reproduction with Qwen3-Coder, ModularSQL preserves the Set-EX baseline (72.06\%) and closes 30.5\% of the reproduced MBS gap, with candidate-free components that transfer unchanged to two additional pipelines' released predictions, adding less than one cent of the LLM cost and 120\,ms amortized latency per 1{,}532-query workload. These results suggest that benchmark execution accuracy does not necessarily imply execution-safe SQL, and that lightweight multiplicity-aware guardrails are a practical complement to existing Text-to-SQL pipelines and evaluation practices. 

\section*{Limitations}
\label{sec:limitations}

\paragraph{Pipeline scope.} The full ModularSQL system, including the candidate-based LLM rescue, is evaluated only on our DeepEye-SQL reproduction. The candidate-free detector and patches transfer to released DAIL-SQL+GPT-4 and BIRD GPT-3.5-turbo predictions (Section~\ref{sec:cross-pipeline}), but these provide only one SQL per question, leaving the rescue stage untested beyond DeepEye-SQL. We found no publicly released BIRD-Dev predictions with candidate pools from other pipelines.

\paragraph{Benchmark and engine.} We evaluate on BIRD-Dev only, on SQLite databases. Spider, BIRD-Test, and production database engines such as PostgreSQL or BigQuery are not yet covered. 

\paragraph{Patch coverage.} The deterministic patch family currently covers only \texttt{DISTINCT}-level rewrites (P1: injection; P2: removal). More complex multiplicity-sensitive operators, including \texttt{GROUP BY} aggregation semantics, nested \texttt{COUNT}/\texttt{SUM} expressions, and JOIN topology rewrites, remain out of scope. Closing the residual 4.31\,pp gap likely requires extending the patch family to address these operators.

\paragraph{LLM rescue prompt.} The rescue prompt does not yet instruct the LLM to prefer candidates that minimize spurious duplicates. Aligning the prompt with the multiplicity-aware probe (e.g., providing each candidate's duplication ratio as part of the context) is a natural low-cost extension.

\paragraph{Threshold calibration.} The P1 duplication-ratio threshold $\tau=0.80$ is calibrated against the P1-fire cases in our DeepEye-SQL reproduction. Other artifact families or query distributions may require re-tuning, although our sensitivity sweep (Appendix~\ref{appx:p1-sensitivity}) shows a plateau over $\tau \in [0.55, 0.80]$. Cross-pipeline transfer (Section~\ref{sec:cross-pipeline}) and fold-level cross-validation (Section~\ref{sec:diagnostics}) provide additional out-of-distribution evidence at the same operating point.

\paragraph{No direct repair comparison.} We do not evaluate an additional repair or regeneration round on flagged queries. The non-LLM selector (Section~\ref{sec:main-results}) isolates the multiplicity-aware signal from generic LLM repair, while direct comparison with execution-guided repair is left to future work.

\bibliography{custom}

\appendix

\section{P1 Threshold Sensitivity}
\label{appx:p1-sensitivity}

The deployed operating point $\tau=0.80$ for P1 (DISTINCT injection) was selected by a sweep on the 87 P1-fire cases in our DeepEye-SQL reproduction (46 fixed + 41 broken at $\tau=0$).  Table~\ref{tab:p1-sweep} reports fix-break counts at representative thresholds. The deployed operating point $\tau=0.80$ sits at the top of a plateau spanning $\tau \in [0.55, 0.80]$, with Net at least $+20$ throughout this range. Leave-one-database-out cross-validation over the 9 databases with P1-fire cases yields a held-out net of $+17$ at the same operating point, indicating stability of the threshold choice beyond the fit in the sample. The protocol difference is worth noting: Table~\ref{tab:p1-sweep} reports a $+22$ gain on the 87-case threshold-selection subset, whereas the end-to-end gain is $+21$ (Table~\ref{tab:main-results}), because the refined rule also triggers on one case outside the calibration subset, causing a PASS-to-FAIL regression.

\begin{table}[!h]
\centering
\scriptsize
\setlength{\tabcolsep}{14pt}
\begin{tabular}{cccc}
\toprule
$\tau$ & Fix & Break & Net \\
\midrule
0.10 & 45 & 34 & $+11$ \\
0.30 & 44 & 26 & $+18$ \\
0.60 & 36 & 15 & $+21$ \\
\rowcolor{blue!10} \textbf{0.80} & \textbf{30} & \textbf{8} & $\mathbf{+22}$ \\
0.90 & 20 & 3 & $+17$ \\
\bottomrule
\end{tabular}
\caption{P1 (DISTINCT injection) threshold sensitivity in our DeepEye-SQL reproduction. Fix and Break count the rule's effect on the 87 P1-fire cases at each $\tau$; Net is $\text{Fix}-\text{Break}$. The deployed operating point $\tau=0.80$ at the top of the plateau is highlighted.}
\label{tab:p1-sweep}
\end{table}

\section{P2 Threshold Sensitivity and Asymmetric Signal}
\label{appx:p2-sensitivity}

We deploy P2 (DISTINCT removal) at the conservative operating point $\tau=0.10$, where the rule fires 5 times in our reproduction with 3 fixes and 0 regressions. Table~\ref{tab:p2-sweep} reports a sweep over four additional thresholds.

\begin{table}[!h]
\centering
\scriptsize
\setlength{\tabcolsep}{14pt}
\begin{tabular}{ccccc}
\toprule
$\tau$ & Fires & Fix & Break & Net \\
\midrule
\rowcolor{blue!10} \textbf{0.10} & \textbf{5} &\textbf{3} & \textbf{0} & $\mathbf{+3}$ \\
0.20 & 8 & 4 & 1 & $+3$ \\
0.30 & 12 & 4 & 4 & $0$ \\
0.50 & 25 & 9 & 9 & $0$ \\
0.80 & 45 & 15 & 18 & $-3$ \\
\bottomrule
\end{tabular}
\caption{P2 (DISTINCT removal) threshold sensitivity in our DeepEye-SQL reproduction. Higher $\tau$ raises the firing rate but does not improve net rescue; the fix-to-break ratio collapses to $1{:}1$ at $\tau=0.30$ and inverts above $\tau=0.50$.}
\label{tab:p2-sweep}
\end{table}

\paragraph{Asymmetric signal.} The duplication ratio threshold $\tau$ behaves asymmetrically across the two patches. P1 (injection) remains a strong signal up to $\tau \approx 0.85$ before its yield collapses; P2 (removal) is actionable only in the extreme tail $\tau \leq 0.20$, with the fix-break tradeoff degrading sharply above this point. This asymmetry suggests that unnecessary \texttt{DISTINCT}s introduced by an LLM-driven pipeline reflect query-intent misclassification rather than mechanical duplicate amplification, and that "add DISTINCT" and "remove DISTINCT" should be treated as independent rules rather than mirror images of one another. This motivates designing the patch family asymmetrically rather than as paired add/remove rules calibrated to a single split point on the duplication ratio. 

\section{LLM Rescue Prompt}
\label{appx:rescue-prompt}

The LLM rescue stage of Algorithm~\ref{alg:rescue} is implemented as a single judge-style query to the backbone model. The exact prompt template used in our experiments is shown below. Placeholders such as \texttt{\{question\}}, \texttt{\{schema\_summary\}}, and \texttt{\{cand\_text\}} are instantiated on a per-query basis. The \texttt{\{cand\_text\}} placeholder expands to twelve candidate entries of the form "\texttt{[i]} SQL: <candidate SQL> Result: <execution preview>", one per candidate. 

\begin{quote}
\small
\ttfamily
You are selecting the best SQL from candidate solutions for a BIRD benchmark question.\\[4pt]
Question: \{question\}\\
Hint: \{hint\}\\[4pt]
Schema (relevant tables/columns):\\
\{schema\_summary\}\\[4pt]
The currently selected SQL has a physical issue (\{trigger\_reason\}):\\
\{base\_sql\}\\
Result: \{base\_preview\}\\[4pt]
Here are 12 alternative candidates with their execution results:\\[4pt]
\{cand\_text\} \\
Pick the candidate that best answers the question. Respond with ONLY the candidate number (0-11), nothing else.\\[4pt]
Your answer:
\end{quote}

\paragraph{Parameters.} We call the model with \texttt{temperature=0.0} and \texttt{max\_tokens=64}. The response is parsed by extracting the first integer in $[0,11]$; if no valid integer is found, the rescue invokes the fallback branch of Algorithm~\ref{alg:rescue} and returns the original S7 selection. Each candidate's SQL is truncated to 600 characters and the schema summary to 1{,}500 characters to keep the prompt within typical context budgets. The prompt contains no multiplicity-aware instructions and does not expose duplication-ratio statistics to the model. Consistent with prior LLM-as-a-judge work \citep{zheng2023llmjudge}, the rescue stage uses \texttt{temperature=0.0} to maximize selection stability and reproducibility. 

\paragraph{Trigger reason placeholder.} The placeholder is set to one of three values aligned with the probe condition in Section~\ref{sec:runtime-probe}: \texttt{cartesian\_explosion} (condition iii: duplication ratio above $\tau$), \texttt{empty\_result} (condition ii), or \texttt{exec\_error} (condition i).  

\section{P2 Implementation Safeguards}
\label{appx:p2-guards}

The P2 rewrite is applied only when all of the following safeguards are satisfied:
\begin{itemize}
    \item The query contains no aggregate function (\texttt{COUNT}, \texttt{SUM}, \texttt{AVG}, etc.).
    \item The query contains no \texttt{GROUP BY} clause.
    \item Removing \texttt{DISTINCT} strictly increases the row count of the executed result.
\end{itemize}

\section{Intervention-Order Ablation: v1 vs v2}
\label{appx:v1v2}

The two intervention orderings introduced in Section~\ref{sec:runtime-probe} differ by 9 queries in net Multiset-EX gain ($+20$ for v1, $+29$ for v2). This appendix isolates where the 9-query gap originates. Rows 4 and 5 of Table~\ref{tab:v1v2-marginal} correspond to the two orderings reported in Table~\ref{tab:main-results}.

\begin{table}[!h]
    \centering
    \scriptsize
    \setlength{\tabcolsep}{10pt}
    \begin{tabular}{lcc}
    \toprule
         Pipeline stage & Multiset-EX & $\Delta$ vs P1+P2  \\
         \midrule
         DeepEye reproduction   & 1009 & --- \\
         + P1 & 1030 & --- \\
         + P1 + P2 & 1033 & $0$ \\
         \midrule
         v1 (P1+P2 $\rightarrow$ Rescue) & 1029 & $-4$ \\
         v2 (Rescue $\rightarrow$ P1+P2) & 1038 & $+5$ \\
         \bottomrule
    \end{tabular}
    \caption{Per-stage Multiset-EX counts (N=1{,}532). The marginal $\Delta$ column reports the gain or loss relative to the +P1+P2 configuration. In v1 the rescue acts on patched SQL and results in a net loss of 4 Multiset-EX-correct queries; in v2 the rescue acts on the original S7 selection before patches refine the result, yielding 5 additional queries beyond patches alone.}
    \label{tab:v1v2-marginal}
\end{table}
\paragraph{Overwrite versus refinement.} In v1, P1 and P2 have already corrected the multiplicity behavior of some flagged queries; the LLM rescue replaces these corrected SQLs with its own selection and in 4 cases, overwrites a patched fix with a worse candidate. In v2, the rescue acts on the original S7 selection, and P1 and P2 then operate on the rescued SQL. The 9-query swing between v1 and v2 isolates the cost of letting the LLM rescue overwrite already-patched SQL. 

\section{By-Difficulty Per-Stage Breakdown}
\label{appx:by-difficulty}

We extend the by-difficulty summary of Table~\ref{tab:by-difficulty} to show the Multiset-EX count at each intermediate pipeline stage. The deterministic patches and the two intervention orderings can then be inspected stratum by stratum.

\begin{table}[!h]
    \centering
    \scriptsize
    \setlength{\tabcolsep}{6pt}
    \begin{tabular}{lcccccc}
    \toprule
         Difficulty & $n$ & Base & + P1 & + P1+P2 & v1 & v2 \\
        \midrule
        Simple & 925  & 662  & 671  & 673  & 671  & 677 \\
        Moderate & 463  & 280  & 287  & 288  & 286  & 288 \\
        Challenging & 144  & 67   & 72   & 72   & 72 & 73 \\
        \midrule
        Total & 1532 & 1009 & 1030 & 1033 & 1029 & 1038 \\
        \bottomrule
    \end{tabular}
    \caption{Multiset-EX counts per BIRD-Dev difficulty stratum at each pipeline stage. "v1" denotes the P1+P2 $\rightarrow$ Rescue ordering and "v2" the canonical Rescue $\rightarrow$ P1+P2 ordering, matching Table~\ref{tab:main-results}.}
    \label{tab:by-difficulty-stages}
\end{table}

\paragraph{Stratum-level v1 vs v2 differences.} The 9-query swing between v1 and v2 ($1038-1029$) splits as $+6$ in the simple stratum, $+2$ in moderate, and $+1$ in challenging. P2 contributes at most two additional fixes within any individual difficulty stratum, consistent with its conservative operating point ($\tau=0.10$) and the asymmetric-signal finding in Appendix~\ref{appx:p2-sensitivity}.

\section{Multiset Equality Implementation Check}
\label{appx:sanity}

Multiset equality between predicted and gold result rows admits two natural implementations: comparison of \texttt{collections.Counter} objects, and string-keyed sorted-tuple equality (\texttt{sorted(rows, key=str)}). On the released Qwen3-Coder DeepEye-SQL prediction artifact, both implementations agree on all $N=1{,}532$ evaluable samples (1019 passes; 0 discrepancies), confirming that the reported Multiset-EX scores are invariant across the two implementations considered here.

\section{Cross-Pipeline Details}
\label{appx:cross-pipeline}
Table~\ref{tab:xpipe-detail} extends the cross-pipeline summary in Section~\ref{sec:cross-pipeline} with per-trigger-reason flagged counts and the P1 patch fix/break breakdown for each system.

\begin{table}[!h]
\centering
\scriptsize
\setlength{\tabcolsep}{3pt}
\resizebox{\columnwidth}{!}{%
\begin{tabular}{lcccccccc}
\toprule
 & \multicolumn{4}{c}{Flagged by reason} & & \multicolumn{3}{c}{P1 patch (Mset-EX)} \\
\cmidrule{2-5} \cmidrule{7-9}
System & Exec & Empty & Cart & Total & P1 & Fix & Brk & Net \\
\midrule
DeepEye-SQL & 1 & 6 & 70 & 77 & 56 & 30 & 8 & $+22$ \\
DAIL-SQL+GPT-4 & 96 & 100 & 74 & 270 & 49 & 22 & 12 & $+10$ \\
GPT-3.5-turbo & 354 & 160 & 41 & 555 & 11 & 8 & 0 & $+8$ \\
\bottomrule
\end{tabular}%
}
\caption{Per-system flagged counts by trigger reason and P1 patch outcomes on BIRD-Dev ($N=1{,}532$). ``Exec'' = execution error, ``Empty'' = empty result set, ``Cart'' = cartesian explosion (dup-ratio $\geq \tau$). P1 fires only on cartesian-flagged queries that do not contain a \texttt{LIMIT} clause. Fix and Brk count Multiset-EX transitions ($0 \!\to\! 1$ and $1 \!\to\! 0$, respectively) among the P1-fired subset.}
\label{tab:xpipe-detail}
\end{table}

\paragraph{Trigger-reason distribution.}
The trigger distribution varies markedly across systems. DeepEye-SQL, the strongest baseline, has few execution errors or empty results (7 combined), whereas GPT-3.5-turbo has 514. Cartesian explosions are more stable (41--74), suggesting that duplicate-generating joins reflect benchmark schema structure rather than model-specific failures.
\paragraph{P1 firing rate.}
P1 fires on 56 of DeepEye-SQL’s 70 cartesian-flagged queries; the 14 excluded cases contain a \texttt{LIMIT} clause. On weaker systems, fewer cartesian queries and the LIMIT guard further reduce the P1-eligible pool (49 for DAIL-SQL, 11 for GPT-3.5-turbo). Despite this smaller pool, the patch maintains a non-negative Multiset-EX net across all three systems, with no Set-EX regressions on DAIL-SQL or GPT-3.5-turbo.
\begin{table}[!h]
\centering
\scriptsize
\setlength{\tabcolsep}{1pt}
\begin{tabular}{lcccc}
\toprule
System & Base Set-EX & New Set-EX & Base Mset-EX & New Mset-EX \\
\midrule
DAIL-SQL+GPT-4 & 53.59\% & 53.59\% & 48.37\% & 49.02\% \\
GPT-3.5-turbo  & 35.90\% & 35.90\% & 32.51\% & 33.03\% \\
\bottomrule
\end{tabular}
\caption{Set-EX and Multiset-EX accuracy before and after the candidate-free guardrail (P1 only) on the two cross-pipeline systems. Set-EX is preserved exactly; Multiset-EX improves by 0.65\,pp and 0.52\,pp, respectively.}
\label{tab:xpipe-accuracy}
\end{table}

\section{Detector Diagnostics Tables}
\label{appx:diagnostics}
This appendix expands the detector diagnostics in Section~\ref{sec:diagnostics} with full precision, recall, and per-trigger breakdowns across all three evaluated systems.

\paragraph{Overall precision and MBS recall.} Table~\ref{tab:detector-pr} reports overall detector precision (the fraction of flagged queries that are true Multiset-EX failures) and MBS recall (the fraction of multiplicity-blind-spot failures that are flagged). The detector flags queries with an execution error (condition~i), an empty result set (condition~ii), or a duplication ratio $\geq \tau$ (condition~iii).

\begin{table}[!h]
\centering
\scriptsize
\setlength{\tabcolsep}{3pt}
\begin{tabular}{lccccccc}
\toprule
 & & \multicolumn{3}{c}{Precision} & \multicolumn{3}{c}{MBS recall} \\
\cmidrule{3-5} \cmidrule{6-8}
System & Flagged & TP & FP & $P$ & MBS & Flagged & $R$ \\
\midrule
DeepEye-SQL & 77 & 65 & 12 & 0.844 & 95 & 36 & 0.379 \\
DAIL-SQL+GPT-4 & 271 & 256 & 15 & 0.945 & 80 & 34 & 0.425 \\
GPT-3.5-turbo & 555 & 550 & 5 & 0.991 & 52 & 18 & 0.346 \\
\bottomrule
\end{tabular}
\caption{Detector precision ($P$) and MBS recall ($R$) on BIRD-Dev ($N=1{,}532$). TP denotes flagged queries that are Multiset-EX failures; FP denotes flagged queries that are Multiset-EX correct. MBS failures pass Set-EX but fail Multiset-EX; the detector prioritizes precision rather than maximizing MBS recall.}
\label{tab:detector-pr}
\end{table}

\paragraph{Per-reason precision.}
Table~\ref{tab:per-reason-precision} reports precision by trigger reason. Execution errors and empty results achieve perfect precision (1.000) across all three systems, as both are definitive failure signals. Cartesian explosion is the only source of false positives, with precision ranging from 0.797 to 0.878.
\begin{table}[!h]
\centering
\scriptsize
\setlength{\tabcolsep}{3pt}
\begin{tabular}{lccc}
\toprule
Trigger reason & DeepEye-SQL & DAIL-SQL+GPT-4 & GPT-3.5-turbo \\
\midrule
\texttt{exec\_error}    & 1.000 (1/1)   & 1.000 (97/97)   & 1.000 (354/354) \\
\texttt{empty\_result}   & 1.000 (6/6)   & 1.000 (100/100) & 1.000 (160/160) \\
\texttt{cartesian\_expl} & 0.829 (58/70) & 0.797 (59/74)   & 0.878 (36/41) \\
\bottomrule
\end{tabular}
\caption{Per-trigger-reason precision. Each cell shows precision followed by (TP/total) for that trigger reason. All false positives originate from the cartesian-explosion trigger.}
\label{tab:per-reason-precision}
\end{table}

\paragraph{Design rationale.}
The detector prioritizes precision over recall: a false positive incurs one LLM rescue call and added latency, but typically leaves the output unchanged if the original candidate is re-selected; a false negative is simply a missed opportunity. The resulting MBS recall (0.35--0.43) reflects this trade-off. The detector catches MBS failures only when they coincide with a runtime anomaly; “silent” MBS failures with normal, non-empty, non-exploded outputs are invisible by construction. Higher MBS recall would require semantic query analysis (e.g., determining whether \texttt{DISTINCT} is warranted), which we leave to future work.

\end{document}